\documentclass[sigconf,nonacm,balance=false]{acmart}

\usepackage{booktabs}
\usepackage{amsmath}
\usepackage{graphicx}
\usepackage{array}
\usepackage{enumitem}
\definecolor{darkblue}{rgb}{0, 0, 0.5}
\hypersetup{colorlinks=true, citecolor=darkblue, linkcolor=darkblue, urlcolor=darkblue}
\begin{document}

\title{Evidence Before Sampling: Interpretable Implicit Negative Candidate Discovery for Recommendation}

\author{\texorpdfstring{Shreya Rajpal\textsuperscript{1*}\quad\allowbreak
  Sonia Sharma\textsuperscript{2}\quad\allowbreak
  Swapnil Parekh\textsuperscript{2}\quad\allowbreak
  Lisa Li\textsuperscript{2}\quad\allowbreak
  Jeyendran Balakrishnan\textsuperscript{2}\quad\allowbreak
  Nagaraj Janardhana\textsuperscript{2}\quad\allowbreak
  Andrew Mattarella-Micke\textsuperscript{2}}{Shreya Rajpal; Sonia Sharma;
  Swapnil Parekh; Lisa Li; Jeyendran Balakrishnan; Nagaraj Janardhana;
  Andrew Mattarella-Micke}}
\affiliation{
  \institution{\textsuperscript{1}Michigan State University, USA\qquad
    \textsuperscript{2}Intuit, USA}
  \city{}
  \country{}
}
\email{rajpalsh@msu.edu}
\email{sonia_sharma@intuit.com}
\email{swapnil_parekh@intuit.com}

\renewcommand{\shortauthors}{Rajpal et al.}

\begin{abstract}
Recommender systems learn from observed user--item interactions, but explicit negative feedback is often unavailable. Since deep learning models require negative signals for training, negative sampling methods typically treat selected unobserved interactions as negatives. However, a missing interaction does not explain why a user is uninterested in an item or whether there is sufficient evidence to label it negative. This is especially important in business recommendation, where negative signals should be interpretable and aligned with business objectives.

We formulate \emph{implicit negative candidate discovery} to identify unobserved interactions supported by observed customer behavior. We encode these patterns as symbolic rules, score them based on support, informativeness, and product relevance, and rank the retained rules by evidence. An LLM then interprets the retained rules using business objectives and domain knowledge; the interpretations are combined with the statistical evidence in the final report.

We evaluate our method in an industrial B2B setting and across five public recommendation datasets. Candidate-quality evaluations in the industrial setting and three public datasets show higher precision than the evaluated baselines, while symbolic selection improves downstream test PR-AUC by 12.5\% over random selection with four negatives per positive example in the industrial task. Our results show that negative candidate validity can be evaluated separately from downstream recommendation performance. This distinction enables evidence-based, business-aligned, and explainable negative selection, improving both interpretability and model training in sparse, skewed, real-world recommendation settings.
\end{abstract}



\maketitle
\begingroup
\renewcommand{\thefootnote}{*}
\begin{NoHyper}
\footnotetext{This work was done as part of an internship at Intuit.}
\end{NoHyper}
\endgroup

\section{Introduction}
\label{sec}
Recommender systems learn user preferences from observed interactions with
items or products, such as clicks and purchases
\cite{rendle2021item}. We refer to these observations as
\emph{user--item interactions}. In real-world data, interactions are observed
for only a small subset of available items. For the rest, a missing
interaction may reflect lack of exposure or missing feedback rather than lack
of interest
\cite{saito2020unbiased,chen2023revisiting,chen2019samwalker}.
Observed clicks or purchases can serve as \emph{positive examples} when
training a recommender. Since explicit negative feedback is often unavailable,
\emph{negative sampling} selects unobserved interactions as negative examples
\cite{rendle2009bpr,ma2026negative}. However, an unobserved interaction may
still correspond to a relevant item, creating a \emph{false negative} and an
incorrect training signal. Existing methods address this through confidence
weights, exposure estimates, or recommender predictions
\cite{chen2020fawmf,liang2016exposure,ding2019rns,ding2020srns,lai2024ahns}.
In real-world business settings, however, statistical or model-based signals
alone may not establish whether a negative candidate is meaningful for a
specific product. The same customer behavior can have different implications
across products and business objectives. We therefore ask a separate question:
\emph{which unobserved interactions have sufficient evidence to be considered
plausible negative candidates, and is that evidence consistent with available
domain knowledge?}
We formulate this problem as \emph{implicit negative candidate discovery}. We
represent recurring customer behaviors as symbolic rules for explainability and evaluate whether
they are sufficiently supported, add information beyond simpler patterns, and
are relevant to the target product. Here, we separate candidate discovery from the
downstream negative sampling task. Once candidates are identified, a sampling strategy can
still decide how many negatives to use and how often to sample them. Our goal
is to establish and record the evidence for a candidate before it is used for
training.\footnote{Code for the public-dataset experiments is available at
\url{https://github.com/Shreyarajpal12/symbolic-sampling}; the industrial data and
pipeline are proprietary.}
We study this problem in a real-world business-to-business (B2B)
customer-support setting with nine products and 10K customer contacts. A
contact corresponds to one customer-support conversation
\cite{sharma2026gradient}. When a product is offered, we observe whether the
customer purchases it within 30 days. Purchases provide positive outcomes,
non-purchases provide observed negative outcomes, and products that were not
offered remain unobserved. We use these observed outcomes to discover symbolic
rules that support negative candidates for unobserved product responses.
A low purchase rate alone is not sufficient evidence for a rule. It may be
based on too few observations, repeat information already captured by a
simpler rule, or reflect generally low purchasing rather than behavior
specific to the target product. We therefore evaluate rules using four
criteria: \textbf{Scope}, \textbf{Evidence Strength},
\textbf{Hierarchical Contribution}, and \textbf{Product Relevance}, which
capture support, uncertainty, added information, and product specificity.
These statistical criteria establish whether a rule is supported by the data,
but not why it should matter for the product. We therefore perform a separate
domain-knowledge review. An LLM receives each retained rule together with the
business objective and trusted domain knowledge supplied by data scientists,
including feature definitions and product descriptions. It does not construct
or score the rules. Instead, it assesses whether the rule has a plausible
interpretation in the supplied context and flags cases without a clear
explanation for human review.
The discovered rules are also useful for data analysis. Data professionals
spend a substantial fraction of their time preparing, inspecting, and
understanding data, with prior studies estimating this effort at roughly
40--45\%
\cite{kaggle2018survey,anaconda2020state,king2015salary,kandel2012enterprise}.
We therefore organize retained rules, statistical evidence, relevant domain
knowledge, and LLM interpretations into a domain-knowledge report. The report
helps data scientists inspect customer patterns and review the evidence behind
negative candidates.
We evaluate candidate quality separately from downstream recommendation
performance. Candidate quality is measured against held-out observed purchase
outcomes and compared with uniform, popularity-based, and matched random
selection, with Bayesian selection baselines additionally evaluated on the public
datasets. In the industrial setting, retained candidates provide a 7.5\%
relative precision gain over the evaluated baselines, and the full statistical
criteria achieve 94.9\% precision. When used for the downstream training task, symbolic selection improves test PR-AUC by
12.5\% relative to random selection.
Negative sampling is therefore one use of the discovered candidates. The same
evidence can also support data analysis and human review while keeping
candidate validity separate from downstream recommendation performance.
\subsection{Contributions}
This paper makes three main contributions:
\begin{itemize}[leftmargin=*]

\item We formulate \emph{implicit negative candidate discovery} as the
problem of identifying unobserved interactions for which observed customer
behavior provides interpretable evidence of a negative outcome, separating
candidate validity from the downstream sampling policy.

\item We evaluate symbolic rules using \textbf{Scope},
\textbf{Evidence Strength}, \textbf{Hierarchical Contribution}, and
\textbf{Product Relevance}. A separate LLM-based domain review relates
retained rules to the business objective and trusted domain knowledge,
records supported interpretations, and flags unclear cases for human
review. These results are summarized in a domain-knowledge report.

\item Symbolic selection achieves higher candidate precision than the evaluated
baselines in the industrial setting and on three public candidate-quality
datasets, with evaluation across five public recommendation datasets overall.

\end{itemize}
\section{Related Work}
\label{sec}

\subsection{Negative Sampling}

Negative-sampling methods differ in how they define useful training examples.
Uniform and popularity-based sampling use fixed distributions, while Dynamic
Negative Item Sampling (DNS) uses the recommender's current scores to select
harder negatives \cite{zhang2013dns}. Adaptive Hardness Negative Sampling
(AHNS) further adjusts negative hardness during training
\cite{lai2024ahns}, while SRNS uses prediction behavior across epochs to reduce
false-negative selection \cite{ding2020srns}. These methods determine
reliability from the recommender and its training behavior on incomplete
interaction data. This makes them well suited to improving optimization, but
the evidence used to select a negative remains model-derived rather than
independently grounded in observed outcomes.

Bayesian Negative Sampling (BNS) addresses false-negative risk by estimating
the posterior probability that an unlabeled interaction is a true negative
using the relative ranking of its recommender-predicted score together with a
false-negative prior \cite{liu2023bns}. This probability is then combined with
sample informativeness to guide negative selection. Although the formulation is
model-agnostic across recommender architectures, its reliability signal still
depends on recommender predictions and interaction statistics. In sparse and
skewed business data, these signals may not capture whether a candidate is
supported by sufficient observations, whether its apparent negative effect is
meaningful relative to the product's baseline purchase rate, whether it adds
information beyond simpler customer patterns, or whether the effect is
specific to the target product.

Recent work also introduces richer learned signals. LLMHNI uses LLM-encoded
semantic relevance and LLM-inferred logical relevance to distinguish hard and
noisy interactions \cite{song2026llmhni}. Since these inferred interactions
can themselves be unreliable, LLMHNI additionally uses graph contrastive
learning to mitigate unreliable edges. Its evidence therefore partly
originates from an LLM and is subsequently refined through learned denoising.
Relying on LLMs themselves to construct or justify evidence can also introduce
faithfulness concerns
\cite{shailya2025lext,turpin2023unfaithful,lyu2024faithful}. We therefore
ground the statistical evidence in observed outcomes before introducing the
LLM.

Positive-Unlabeled Recommendation Learning (PURL) takes a different approach
and learns directly from positive and unlabeled feedback without requiring
sampled negative labels \cite{wang2025purl}. This is useful when reliable
negative labels are unavailable. However, in real-world settings where
observed non-purchase outcomes are abundant and may substantially outnumber
positive outcomes, avoiding explicit negative supervision can leave useful
behavioral information unused. Evidence-backed negative discovery instead
selects such signals based on observed support rather than treating all
unobserved interactions as negative or avoiding negative candidates
altogether.

Building on these efforts, we formulate \emph{implicit negative candidate
discovery} as a step before negative sampling. Candidate reliability is
decomposed into four complementary criteria rather than represented by a single
model-derived, posterior, or semantic signal. \textbf{Scope} measures whether
the pattern has sufficient observable support. \textbf{Evidence Strength}
measures whether its product-relative reduction in purchase remains credible
under finite-sample uncertainty. \textbf{Hierarchical Contribution} measures
whether a more specific condition adds information beyond simpler conditions.
\textbf{Product Relevance} measures whether the effect is specific to the
target product rather than reflecting general low-purchasing behavior. These
criteria correspond to different failure modes and are applied as separate
gates, so evidence on one dimension cannot compensate for failure on another.
The LLM is introduced only after these statistical checks and relates retained
rules to trusted domain knowledge and the business objective. This separates
measured evidence from generated interpretation. The retained candidates can
then be used by downstream sampling strategies.

\subsection{Symbolic Methods}

Our rule construction is related to subgroup and pattern discovery, where
symbolic descriptions identify populations whose outcomes differ from a
reference group
\cite{kraljnovak2009supervised,komiyama2017statistical}. Closed-pattern mining
reduces descriptions that correspond to the same observed population
\cite{zaki2002charm}. These methods primarily identify distinctive or
nonredundant subgroups. Our symbolic rules additionally test whether a more
specific pattern contributes beyond its simpler conditions and whether that
effect is specific to the target product. This distinguishes product-related
evidence from redundant patterns and general low-purchasing behavior.

Symbolic representations have also been used to make LLM reasoning more
explicit, structured, and inspectable across domains
\cite{gao2023pal,pan2023logiclm,chen2021finqa,rajpal2026spatial}. These works
show that explicit intermediate representations can provide a more
interpretable basis for reasoning than unconstrained language-model inference.
Here, symbolic rules provide explicit evidence that can subsequently be
interpreted using product and domain knowledge.

Our formulation is also related to weak supervision, where rules or imperfect
labeling functions construct training labels
\cite{ratner2017snorkel,varma2018snuba}. Weak-supervision rules act as labeling
sources whose outputs are combined to infer training labels. Our rules are
instead evaluated using statistical evidence from observed purchase outcomes
before they are used to identify candidate negatives. This preserves the
distinction between an observed behavioral pattern and the unknown preference
associated with an unobserved interaction.
\section{Methodology}
\label{sec:method}

\subsection{Overview}

\textbf{Problem Definition:} We use observed customer features and purchase
outcomes to identify conditions under which customers purchase a product less
often. We express each condition as a rule, such as low account balance
together with low direct deposit. For a product $p$, we evaluate the rule using
customers whose outcome for $p$ is observed. We then use a supported rule to
identify negative candidates among customers whose response to $p$ is
unobserved. For the formal description, a customer and a target product are
written as a customer--product pair $(u,p)$. The pair identifies the outcome
we want to assess; it does not imply that an interaction was observed.

We evaluate each rule separately for each product. This is necessary because
a rule can be supported by too few observations, repeat a simpler condition,
or describe low purchasing across products. The scores show which of these
checks the rule passes. We rank retained rules within each product using their
evidence scores. During LLM inference, the model relates the rules to the
business objective and domain knowledge supplied by data scientists. We report
the ranked rules, scores, and interpretations together. The report gives data
scientists a starting point for examining product-related behavior and aims to
reduce part of the manual exploratory work required before downstream modeling.

\begin{figure*}[t]
    \centering
    \includegraphics[width=\textwidth]{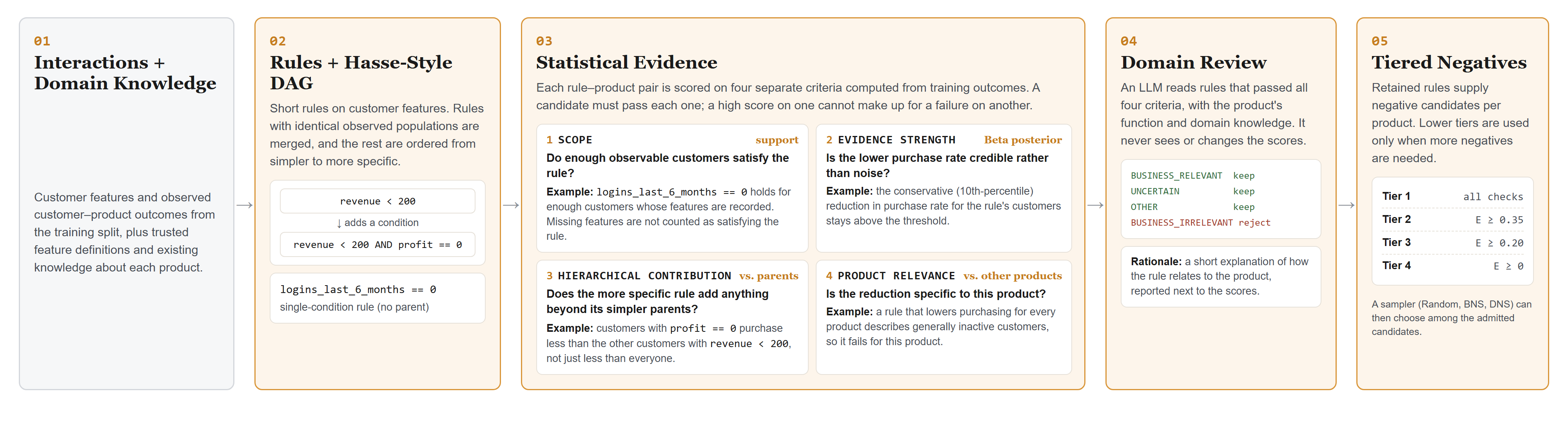}
    \caption{The symbolic discovery procedure. Rules are evaluated using
    Scope, Evidence Strength, Hierarchical Contribution, and Product Relevance,
    followed by a separate domain review.}
    \Description{Pipeline from observed data to symbolic rules, followed by
    Scope, Evidence Strength, Hierarchical Contribution, Product Relevance,
    and a separate domain review.}
    \label{fig:method-overview}
\end{figure*}
Figure~\ref{fig:method-overview} shows the procedure. We first construct short
rules from customer features. We merge rules that identify the same observed
customers, then organize the remaining rules from simpler to more specific
conditions. We evaluate each rule--product pair using Scope, Evidence Strength,
Hierarchical Contribution, and Product Relevance. Rules that pass these checks
are reviewed using the business objective and domain knowledge supplied by data
scientists. The retained rules are used to construct negative candidates and a
ranked domain-knowledge report.

Rule construction and statistical estimation use training data. Validation
outcomes select the operating thresholds. Test outcomes are reserved for final
evaluation.

\subsection{Rule Construction}
\label{sec:candidates}

\paragraph{Users, features, and predicates.}

We are given a dataset containing customer features, customer interactions with different products, and product information, together with domain knowledge describing the features and products. Let $u$ denote a user and $x_{uj}$ denote the value of feature $j$ for user $u$. These features describe observed properties of the customer, such as checking balance, direct-deposit amount, account tenure, or existing product usage.

We remove features that are insufficiently observed or nearly constant in the training data. For each remaining numerical feature, we construct simple conditions, called predicates, using the training-set 25th, 50th, and 75th percentiles. These correspond to the lower quartile, median, and upper quartile. We also include a zero-valued predicate when zero has a meaningful interpretation. For example, if the 25th percentile of checking balance is \$500, the predicate $\text{checking balance}\leq \$500$ identifies customers in approximately the lowest quartile of the training population.

We enumerate each individual predicate and each conjunction of two predicates from different feature columns. A conjunction requires both conditions to hold. For instance, a rule may be written as
\[
r:
\quad
\text{checking balance}\leq \$500
\land
\text{direct deposit}\leq \$1{,}200.
\]
Here, $r$ denotes the complete customer rule, while the individual conditions are the predicates from which it is constructed.

\paragraph{Observed and Satisfying Users.}

For a rule $r$, let $\mathcal{J}(r)$ denote the features required to evaluate the rule. We denote the users for whom all features in $\mathcal{J}(r)$ are observed by $\mathcal{U}^{\mathrm{obs}}_r$, and the users who satisfy the rule by $\mathcal{U}^{\mathrm{sat}}_r\subseteq\mathcal{U}^{\mathrm{obs}}_r$. For example, consider the rule
$\text{checking balance}\leq \$500 \land \text{direct deposit}\leq \$1{,}200$.
A user for whom both features are observed, but whose checking balance is above \$500, belongs to $\mathcal{U}^{\mathrm{obs}}_r$ but not to $\mathcal{U}^{\mathrm{sat}}_r$. In contrast, a user with missing direct-deposit information cannot be evaluated for the complete rule and therefore does not belong to $\mathcal{U}^{\mathrm{obs}}_r$.
\subsection{Rule Organization with a Hasse-Style DAG}
\label{sec:states}

Different rules may describe the same or closely related customer populations. We construct rules with at most two predicates. We merge two rules when their tri-state signatures are identical on the training data, meaning that they have the same observable users and the same users satisfying the rule. We organize the remaining rules from simpler to more specific conditions in a directed acyclic graph (DAG). We keep only immediate relationships, giving a Hasse-style DAG as described below. A rule $r_c$ is a \emph{child} of $r_p$ when the customer condition represented by $r_c$ is a more specific version of the condition represented by $r_p$. For example,
$\text{checking balance}\leq \$500$
is a parent of
$\text{checking balance}\leq \$500 \land \text{direct deposit}\leq \$1{,}200$,
since the child retains the parent's condition and introduces an additional restriction. A rule may have multiple parents when it refines more than one simpler condition.

Having all of one rule's users inside another rule's population is not enough to define a parent--child relationship. A broad but unrelated rule may happen to contain all users satisfying a smaller rule and would therefore provide a misleading baseline for evaluating the contribution of the child. We therefore first remove predicates that are shared by the parent and child and examine the conditions that distinguish the two rules. If the remaining condition corresponds to adding a predicate or tightening a threshold on the same feature, the relationship is retained directly. When the remaining conditions involve different features, we retain the relationship only when the features are sufficiently associated in the training data, the child population is strictly contained within the parent population, and the parent contains enough remaining users to support a reliable comparison. The discovery predicates are numerical, so we measure association using Spearman correlation.

For example, consider the rules
$\text{payment app usage rate}=0 \land \text{revenue}\leq \$1{,}590$
and
$\text{payment app usage rate}=0 \land \text{profit}=0$.
After removing the shared payment-app condition, the potential relationship depends on whether
$\text{revenue}\leq \$1{,}590$
provides a meaningful parent condition for
$\text{profit}=0$.
Although revenue and profit are sufficiently associated and the 77 users satisfying the latter rule are contained within the 79 users satisfying the former, only two users remain in the parent after excluding the child. We therefore reject this edge because it would provide too small a comparison population for evaluating the child's additional contribution.

Finally, we retain only immediate parent--child relationships. If a relationship between two rules is already explained through an intermediate rule, the direct edge is removed. This produces a Hasse-style DAG in which each parent represents a meaningful simpler condition against which the additional contribution of a child rule can later be evaluated.

\subsubsection{Scope}
\label{sec:scope}

Scope measures how many customers satisfy a rule and have an observed product outcome. Let $\mathcal{U}^{\mathrm{out}}_p$ denote the users for whom the outcome of product $p$ is known. We define the Scope of rule $r$ for product $p$ as

\begin{equation}
S_{r,p}
=
\frac{
\left|
\mathcal{U}^{\mathrm{sat}}_r
\cap
\mathcal{U}^{\mathrm{out}}_p
\right|
}{
\left|
\mathcal{U}^{\mathrm{out}}_p
\right|
}.
\label{eq:scope}
\end{equation}

Thus, $S_{r,p}$ measures the fraction of users with a known product outcome who also satisfy rule $r$. We denote its numerator by $n^{\mathrm{sat}}_{r,p}$, the number of users with a known outcome for $p$ who satisfy $r$. We also define the unknown fraction as
\begin{equation}
u_{r,p}
=
1-
\frac{
\left|\mathcal{U}^{\mathrm{obs}}_r\cap\mathcal{U}^{\mathrm{out}}_p\right|
}{
\left|\mathcal{U}^{\mathrm{out}}_p\right|
}.
\label{eq:unknown-fraction}
\end{equation}
Because a user can satisfy the rule only when all required features are observed, these quantities capture both support and observability.

For example, suppose the outcome of product $p$ is known for 1{,}000 users. The rule
$\text{checking balance}\leq \$500 \land \text{direct deposit}\leq \$1{,}200$
can be evaluated for 900 of these users, and 180 users satisfy it. The resulting Scope is $180/1000=0.18$, meaning that the rule is supported by 18\% of the customer population for which the product outcome is known.

Scope describes the fraction of customers supporting a rule. The fraction alone
does not establish that the sample is large enough, so we also require a
minimum number of observed outcomes. Equation~\ref{eq:candidate-scope-gates}
gives the support and missing-feature requirements. We select their thresholds
using observed validation outcomes. Missing product outcomes do not count as
evidence of non-purchase.

\subsubsection{Evidence Strength}
\label{sec:evidence}

Evidence Strength measures the reduction in purchase rate among users satisfying a rule. It accounts for uncertainty in the estimate because the observed rate depends on a finite number of users.

For a rule--product pair $(r,p)$, let $n_{r,p}$ denote the number of users satisfying rule $r$ for whom the outcome of product $p$ is known. Let $c_{r,p}$ denote the number of these users who purchase the product. Their observed purchase rate is

\begin{equation}
q_{r,p}
=
\frac{c_{r,p}}{n_{r,p}}.
\end{equation}

We compare this rate with the purchase rate among all users for whom both the rule features and the product outcome are observed. We denote this baseline by $q_{0,p}$. It includes users who satisfy the rule and those who do not. For a positive baseline rate, the relative purchase reduction is

\begin{equation}
D_{r,p}
=
1-
\frac{q_{r,p}}{q_{0,p}}.
\label{eq:relative-reduction}
\end{equation}

A positive $D_{r,p}$ indicates that users satisfying the rule purchase product $p$ less often than the product baseline. For example, suppose 8 of 180 users satisfying a rule purchase product $p$. This gives $q_{r,p}=8/180=0.044$. Suppose 116 of the 900 evaluable users purchase the product overall. The baseline purchase rate is then $q_{0,p}=116/900=0.129$. This gives a relative purchase reduction of approximately $D_{r,p}=0.66$. In other words, the observed purchase rate among users satisfying the rule is about 66\% lower than the baseline purchase rate.

The baseline matters because some products are rarely purchased across the
population. Comparing $q_{r,p}$ with $q_{0,p}$ measures whether the rule
identifies an additional reduction for that product.

The same observed reduction can provide different evidence depending on the
number of users. We account for this uncertainty in the purchase rates below.

We model the purchase probabilities among users satisfying and not satisfying the rule using a Jeffreys prior, $\operatorname{Beta}(1/2,1/2)$. The comparison population contains only evaluable non-satisfiers:
\[
\mathcal{U}^{\mathrm{neg}}_{r,p}
=
\left(\mathcal{U}^{\mathrm{obs}}_r
\setminus
\mathcal{U}^{\mathrm{sat}}_r\right)
\cap
\mathcal{U}^{\mathrm{out}}_p.
\]
Let $n_{\neg r,p}=|\mathcal{U}^{\mathrm{neg}}_{r,p}|$ and let $c_{\neg r,p}$ denote the number of purchases in this set. Thus, customers with a missing feature required by $r$ are excluded from both groups rather than treated as non-satisfiers. The posterior purchase probabilities are

\begin{equation}
q_{r,p}
\mid c_{r,p},n_{r,p}
\sim
\operatorname{Beta}
\left(
c_{r,p}+\frac{1}{2},
n_{r,p}-c_{r,p}+\frac{1}{2}
\right),
\end{equation}

and

\begin{equation}
q_{\neg r,p}
\mid c_{\neg r,p},n_{\neg r,p}
\sim
\operatorname{Beta}
\left(
c_{\neg r,p}+\frac{1}{2},
n_{\neg r,p}-c_{\neg r,p}+\frac{1}{2}
\right).
\end{equation}

These posterior distributions represent uncertainty in the purchase probabilities given the purchase and non-purchase counts. They let us calculate a range of purchase-rate reductions instead of relying on the observed rates alone.

We draw $B$ posterior samples from these distributions. For each draw $b$, we reconstruct the corresponding product baseline as

\begin{equation}
q_{0,p}^{(b)}
=
\frac{
n_{r,p}q_{r,p}^{(b)}
+
n_{\neg r,p}q_{\neg r,p}^{(b)}
}{
n_{r,p}+n_{\neg r,p}
}.
\end{equation}

We then recompute the relative purchase reduction for that draw,

\begin{equation}
D_{r,p}^{(b)}
=
1-
\frac{q_{r,p}^{(b)}}{q_{0,p}^{(b)}}.
\end{equation}

We repeat this process for $B=10{,}000$ draws. This produces a distribution of plausible relative purchase reductions instead of a single observed value. We define Evidence Strength as

\begin{equation}
E_{r,p}
=
\max
\left(
0,\;
Q_{0.10}
\left(
D_{r,p}^{(1:B)}
\right)
\right),
\label{eq:evidence-strength}
\end{equation}

where $Q_{0.10}$ denotes the 10th percentile. For example, if the observed reduction is $0.66$ and the posterior 10th percentile is $0.48$, the Evidence Strength is $0.48$. The outer maximum assigns zero when this percentile does not support a reduction. Thus, $D_{r,p}$ describes the observed reduction, while $E_{r,p}$ accounts for uncertainty in the purchase rates.

We retain rule--product pairs whose Evidence Strength satisfies $E_{r,p}\geq\tau_E$. We
select this threshold using the non-purchase rate of the resulting candidates
on the validation set. The rules and their scores are computed from training
data.

\subsubsection{Relevance}
\label{sec:relevance}

Relevance evaluates whether a rule adds information beyond simpler customer
conditions and whether that information is specific to the target product.
We measure these properties using Hierarchical Contribution and Product
Relevance, then separately review the rule's business meaning.

\paragraph{Hierarchical Contribution.}

A more specific rule should provide information beyond the simpler rules from which it is derived. Let $r_c$ denote a child rule and $r_p$ one of its immediate parents in the Hasse-style DAG. Because all users satisfying $r_c$ also satisfy $r_p$, directly comparing the child with the complete parent population would include the child population in both groups. We therefore compare users satisfying the child with the evaluable parent remainder,
\[
\mathcal{U}^{\mathrm{rem}}_{r_p,r_c,p}
=
\left(
\mathcal{U}^{\mathrm{sat}}_{r_p}
\cap
\mathcal{U}^{\mathrm{obs}}_{r_c}
\cap
\mathcal{U}^{\mathrm{out}}_p
\right)
\setminus
\mathcal{U}^{\mathrm{sat}}_{r_c},
\]
which we denote by $r_p\setminus r_c$ for brevity. Customers who satisfy the parent but have a missing feature required by the child are excluded from this comparison rather than treated as child non-satisfiers.

For example, suppose
$\text{checking balance}\leq \$500$
is satisfied by 300 users, while the more specific rule
$\text{checking balance}\leq \$500 \land \text{direct deposit}\leq \$1{,}200$
is satisfied by 180 of them. If all 300 parent users are evaluable for the child rule, the parent remainder contains the 120 users who have low checking balance but do not satisfy the complete child rule. This comparison isolates whether the additional condition introduced by the child is associated with a further reduction in purchase behavior.

For each posterior draw $b$ and parent--child pair, we define the hierarchical contribution as

\begin{equation}
H^{(b)}_{r_c\mid r_p,p}
=
1-
\frac{q^{(b)}_{r_c,p}}
    {q^{(b)}_{r_p\setminus r_c,p}}.
\label{eq:hierarchical-contribution-parent}
\end{equation}

A rule may have multiple immediate parents, denoted by $\mathrm{pa}(r_c)$.
For each posterior draw, we take the smallest contribution across its parent
comparisons:

\begin{equation}
H^{(b)}_{r_c,p}
=
\min_{r_p\in\mathrm{pa}(r_c)}
H^{(b)}_{r_c\mid r_p,p}.
\label{eq:hierarchical-contribution-draw}
\end{equation}

We summarize these draws conservatively as

\begin{equation}
H_{r_c,p}
=
Q_{0.10}
\left(
H^{(1:B)}_{r_c,p}
\right).
\label{eq:hierarchical-contribution}
\end{equation}

Here, $q^{(b)}_{r_c,p}$ and $q^{(b)}_{r_p\setminus r_c,p}$ are posterior draws of the purchase rates for the child and parent-remainder populations, respectively. A positive $H_{r_c,p}$ means that the 10th percentile of this minimum remains positive. This checks the child against each immediate parent, so a favorable comparison with one parent cannot compensate for a weak comparison with another.

\paragraph{Product Relevance.}

A rule may identify customers who purchase less across several products. We need to check whether its additional condition is more informative for the target product. We therefore compare the hierarchical effect for product $p$ with the corresponding effects observed for other products. For each posterior draw, we define

\begin{equation}
R_{r,p}^{(b)}
=
H_{r,p}^{(b)}
-
\operatorname{median}_{p'\neq p}
H_{r,p'}^{(b)},
\end{equation}

and summarize this difference conservatively as

\begin{equation}
R_{r,p}
=
Q_{0.10}
\left(
R_{r,p}^{(1:B)}
\right).
\end{equation}

A positive value indicates that the added condition is associated with a larger
purchase-rate reduction for the target product than for the median comparison
product. When too few comparison products are available, Product Relevance is
marked as not estimable rather than assigned zero.

\paragraph{Domain-knowledge relevance.}
After a rule passes the statistical checks, an LLM reviews whether the
customer conditions have a plausible connection to the target product.
It receives the symbolic rule, the product's function, trusted feature
definitions, and existing domain knowledge. It does not receive or modify the
statistical scores during this review. For example, low login activity may
coincide with fewer purchases, but its meaning depends on whether that activity
is relevant to how the product is used.

The LLM assigns one of four labels and gives a short rationale.
\texttt{BUSINESS\_RELEVANT} indicates a plausible product-related
interpretation. \texttt{UNCERTAIN} indicates that the available domain
knowledge is insufficient for a confident judgment. \texttt{OTHER} is used
when the rule does not fit a predefined interpretation but is not judged
invalid. These three labels are retained. Only
\texttt{BUSINESS\_IRRELEVANT} rules are rejected. The LLM therefore makes a
qualitative filtering decision after the statistical evaluation. Its rationale
does not add numerical evidence or establish why the observed pattern occurs.

\subsection{Candidate Admission}
\label{sec:selection}

We define the final candidate set using explicit gates rather than combining
Scope, Evidence Strength, Hierarchical Contribution, and Product Relevance into
a single confidence score. These quantities capture different validity
conditions, and a strong value on one dimension cannot compensate for failure
on another.

A rule--product pair is admitted to the statistically supported candidate set
when it satisfies

\begingroup\small
\begin{align}
\textbf{Scope:}\quad
&n^{\mathrm{sat}}_{r,p,\mathrm{train}}\geq n_{\min},
\qquad
u_{r,p,\mathrm{train}}\leq\tau_u,
\label{eq:candidate-scope-gates}
\\
\textbf{Evidence Strength:}\quad
&D_{r,p}>0,
\qquad
E_{r,p}\geq\tau_E,
\label{eq:candidate-evidence-gates}
\\
\textbf{Hierarchical Contribution:}\quad
&H_{r,p}>\tau_H,
\label{eq:candidate-hierarchy-gates}
\\
\textbf{Product Relevance:}\quad
&R_{r,p}>\tau_R.
\label{eq:candidate-relevance-gates}
\end{align}
\endgroup

Here, $n_{\min}$ is the minimum number of training examples satisfying a rule
for which product $p$ can be evaluated, $\tau_u$ is the maximum allowed
unknown fraction, $\tau_E$ is the minimum Evidence Strength,
$\tau_H$ is the hierarchical-contribution threshold, and $\tau_R$ is the
product-specific relevance threshold. The gates are applied sequentially so
that each retained rule is sufficiently observed, exhibits credible negative
evidence, contributes beyond its simpler explanations, and is relevant to the
target product.
The Scope gate in this section describes the industrial run. The public
pipeline uses $S>0$ and does not apply $n_{\min}$ or $\tau_u$.

Some rules can have strong empirical evidence while lacking the information
required for complete relevance estimation. This can occur for root rules,
rules with insufficient parent-remainder support, or cases where suitable
cross-product comparisons are unavailable. We retain such rules separately as
\texttt{SUPPORTED\_\allowbreak UNVERIFIED}; they are not included in the set that passes
all four statistical criteria.

\paragraph{Domain-knowledge synthesis.}
We rank the retained rules within each product using their evidence scores and
produce a Markdown report containing the ranked rules, scores, and proposed
business interpretations. During LLM inference, the model receives the rules,
business objective, and domain knowledge supplied by data scientists. The
statistical scores are combined with the generated interpretations when the
final report is assembled.

This report gives data scientists specific customer conditions to examine when
reviewing product-related behavior. The same rules can be used to select
negative training examples. These are two uses of the rule evidence; a useful
training result does not by itself validate the proposed business explanation.

\subsection{Downstream Tiered Negatives}
\label{sec:tiered-negatives}

We select negative training examples from the retained rules in tiers. Each
tier records which requirements a rule satisfies. \textbf{Tier 1} contains rules that satisfy the full
statistical procedure and pass the domain-knowledge gate, and therefore
contains the rules retained after both statistical evaluation and business
review. When additional negatives
are required, the system progressively descends to lower tiers by relaxing
the Evidence Strength threshold: Tier 2 uses $E\geq0.35$, Tier 3 uses
$E\geq0.20$, and Tier 4 admits rules with $E\geq0$.

The remaining validity requirements are kept fixed across tiers. Rules must
retain sufficient support, the estimated effect must remain in the negative
direction ($D>0$), and a rule with an estimable $R\leq0$ is excluded. The
tiering procedure is applied independently for each product because products
have different numbers of candidates that pass the stricter thresholds. In the downstream
experiments, tiers determine the order in which candidates are selected to fill
the negative budget. The evaluation treats tiering as a candidate-selection
policy rather than an additional loss-weighting mechanism; tier and evidence
information are retained as audit metadata.
The public experiments use the analogous ordered ladder in
Table~\ref{tab:implementation-settings}, which places rules that pass Evidence
Strength but fail Hierarchical Contribution or Product Relevance before the
relaxed Evidence levels.

\section{Evaluation}
\label{sec:evaluation}

\subsection{Experimental Setup}
\label{sec:setup}

\paragraph{Industrial data.}

We use the customer-support setting described by
Sharma et al.~\cite{sharma2026gradient}. A contact is one conversation, and a
chunk is a segment of 20 utterances from that conversation. The source
contains 93,001 chunk-level rows, which are deduplicated to 10,000 customer
contacts for rule discovery. The catalog contains nine B2B products.
Table~\ref{tab:pipeline-stats} summarizes the data and rule counts. The
discovery splits are contact-disjoint and temporally ordered, so the same
contact does not appear in more than one split.

\begin{table}[t]
\centering
\small
\caption{Industrial data and rule counts before downstream recommender training.}
\label{tab:pipeline-stats}
\begin{tabular}{@{}lr@{}}
\toprule
Quantity & Count \\
\midrule
Raw chunk rows & 93,001 \\
Deduplicated contacts & 10,000 \\
Products & 9 \\
Initial feature columns & 121 \\
Usable numeric features & 101 \\
Atomic predicates & 225 \\
Unary/pairwise expressions & 6,336 \\
Distinct empirical rules & 5,117 \\
Frozen Hasse edges & 10,704 \\
\bottomrule
\end{tabular}
\end{table}

\paragraph{Discovery parameters.}
Predicate thresholds are learned from the training split only. Both pipelines
use a Jeffreys $\operatorname{Beta}(0.5,0.5)$ prior, $B=10{,}000$ posterior
draws, and the 10th percentile for $E$, $H$, and $R$. The industrial pipeline
uses $n_{\min}=30$ and $\tau_E=0.45$. Its tier cut-points are 0.35, 0.20, and
0. The public pipeline uses $S>0$, a minimum arm size of 5 for estimable $H$
and $R$ comparisons, and the same $\tau_E=0.45$ without retuning. Discovery
thresholds are selected without using downstream recommender performance.

\begin{table*}[t]
\centering
\scriptsize
\setlength{\tabcolsep}{3pt}
\caption{Implementation settings for the industrial and public pipelines.
The industrial pipeline is proprietary; settings described in words were
fixed on validation data before downstream training.}
\label{tab:implementation-settings}
\begin{tabular}{@{}p{0.18\textwidth}p{0.35\textwidth}p{0.40\textwidth}@{}}
\toprule
\textbf{Component} & \textbf{Industrial pipeline} & \textbf{Public pipeline} \\
\midrule
Rules & Train quartiles and zero predicates; maximum arity 2; identical
tri-state signatures merged & Same, with categorical level predicates;
maximum arity 2; identical tri-state signatures merged \\
Posterior & $\operatorname{Beta}(0.5,0.5)$; 10,000 draws; 10th percentile & Same; posterior seed 42 \\
Scope & $n_{\min}=30$; maximum unknown fraction $\tau_u$ fixed on validation
data before downstream training & $S>0$; no $n_{\min}$ or $\tau_u$ \\
Thresholds & $\tau_E=0.45$, $\tau_H=0$, $\tau_R=0$ &
$\tau_E=0.45$, $\tau_H=0$, $\tau_R=0$ \\
Hasse edges & Same-feature refinements kept directly; cross-feature edges
require Spearman association between the distinguishing features and a
minimum parent remainder, both fixed before downstream training. A minimum
arm size is enforced for $H$ and $R$ estimates (5 in the public pipeline). &
Predicate-extension edges only; minimum child and parent-remainder arm size 5 \\
Fallback tiers & $E\geq0.35$, $E\geq0.20$, $E\geq0$ & Levels 0--6, from
$S+E+H+R$ through uniform random fill \\
Negative budget & $K\in\{2,4,6\}$; BNS uses $K=2$ & $K=4$ \\
Backbone & Proprietary ranking model, identical across all negative-selection methods &
LightGCN (64 dimensions, 2 layers) on KuaiRec; DeepFM (32 dimensions,
256--128 MLP, dropout 0.2) on MIND and RetailRocket \\
Training & Validation Case-Load checkpoint selection &
BCE, Adam, batch size 4,096, no weight decay; learning rate $5\times10^{-3}$
for LightGCN and $5\times10^{-4}$ for DeepFM; at most 40 epochs; validation
PR-AUC early stopping with patience 15; seeds 42, 1, and 2 \\
\bottomrule
\end{tabular}
\end{table*}

For public experiments, LightGCN uses a graph constructed from training
positives. DeepFM uses user, item, category, and user-feature fields. All
public models use pointwise binary cross-entropy and restore the checkpoint
with the highest validation PR-AUC, breaking ties by validation Top-1. The
seed controls model initialization, batch order, and negative sampling.

\paragraph{Downstream model and checkpoint selection.}
Every negative-selection method is evaluated with the same downstream recommender architecture, positive examples, optimizer, maximum epoch budget, validation population, and checkpoint-selection rule. Validation Case-Load performance selects the checkpoint for every method; PR-AUC and the operational metric are then reported at that same checkpoint. Test data are never used for checkpoint selection.

\paragraph{Baselines.}
The downstream experiments compare symbolic selection with uniform random
selection and Bayesian Negative Sampling (BNS) \cite{liu2023bns}.
Random and symbolic selection use matched budgets of two, four, and six
negatives per positive example; BNS is evaluated at two. Industrial
candidate-quality experiments use uniform, popularity-based, and
product-matched random controls. The public-dataset experiments also
include Bayesian selection and use category-matched controls where
appropriate. Candidate counts are matched within each comparison. The
downstream methods receive the same positive training examples.
Public BNS and BNS-S are item-level reimplementations of BNS rather than the
full model-score method.

\paragraph{Metrics.}
We evaluate candidate quality using held-out observed outcomes because we do
not know the customer's response to a product that was never offered.
\emph{Negative-candidate precision} is the fraction of selected candidates
whose observed outcome is negative. We also report coverage, the fraction of
eligible examples selected, and contamination, the number of observed
purchases among those selected examples. Downstream evaluation uses PR-AUC
and three conversation-stage metrics: Case-Load (CL), Early Recommendation
(ER), and Ground Zero (GZ). Following
Sharma et al.~\cite{sharma2026gradient}, these are micro-F1 at the first
chunk, second chunk, and highest-weight chunk for each contact, respectively.
They measure recommendation quality at different points in a conversation.
Higher values are better.

\section{Results}
\label{sec:results}
\subsection{RQ1: How Do Evidence-Guided Negative Candidates Affect Downstream Recommendation?}
\label{sec:rq1-downstream}

Table~\ref{tab:negative-sampling-early-stopping} compares random, symbolic,
and Bayesian negative sampling under Case-Load-based early stopping.
$K$ is the number of unobserved customer--product interactions added as
negative training examples per positive example. $K=0$ uses no additional
implicit negatives. Random and symbolic sampling use matched budgets
$K\in\{2,4,6\}$, while BNS is evaluated at $K=2$. At inference, the
recommender scores each customer's candidate products and ranks them by
predicted purchase likelihood.

\begin{table}[t]
\centering
\caption{
Comparison of random, symbolic, and Bayesian negative-sampling strategies
under Case-Load-based early stopping for different numbers of implicit
negatives $K$, the number of unobserved interactions added as negative
examples per positive training example. $K=0$ uses no additional implicit
negatives. 
CL, ER, and GZ are micro-F1 at the first chunk, second chunk, and highest-weight
chunk, respectively; higher is better.
}
\label{tab:negative-sampling-early-stopping}
\scriptsize
\setlength{\tabcolsep}{2pt}
\begin{tabular}{@{}clrrrrr@{}}
\toprule
$K$ & Sampling Strategy & CL & ER & GZ & Val PR-AUC & Test PR-AUC \\
\midrule

0 & No Implicit Negatives
& 0.3989 & 0.4277 & 0.4708 & 0.7774 & 0.5071 \\

\midrule

2 & Random
& 0.4706 & 0.5181 & 0.4987 & 0.7153 & 0.5907 \\

2 & Symbolic
& 0.4770 & 0.5319 & 0.5061 & 0.7617 & 0.5837 \\

2 & Bayesian
& 0.3841 & 0.3825 & 0.3826 & 0.4790 & 0.3780 \\

\midrule

4 & Random
& 0.4120 & 0.4706 & 0.4541 & 0.6582 & 0.5166 \\

4 & Symbolic
& 0.4562 & 0.5004 & 0.4861 & 0.6866 & 0.5814 \\

\midrule

6 & Random
& 0.3681 & 0.4744 & 0.4398 & 0.6185 & 0.5586 \\

6 & Symbolic
& 0.4302 & 0.5012 & 0.4659 & 0.6906 & 0.5682 \\

\bottomrule
\end{tabular}
\end{table}

\paragraph{Results.}
Symbolic selection has higher CL, ER, and GZ than random selection at each
matched budget. At $K=4$, CL increases from 0.4120 to 0.4562 and test PR-AUC
from 0.5166 to 0.5814. The PR-AUC gain is 12.5\% relative to random selection.
At $K=6$, symbolic selection also has higher test PR-AUC. At $K=2$, however,
random selection gives higher test PR-AUC, 0.5907 compared with 0.5837 for
symbolic selection. BNS performs below the no-additional-negatives baseline
in the reported $K=2$ configuration.

Adding more negatives does not consistently improve the model. Within the
symbolic runs, $K=2$ gives the highest CL, ER, GZ, and test PR-AUC among the
tested budgets. The gain over random selection at $K=4$ therefore does not
mean that four negatives is the best budget. These results show that the
selection method and the number of negatives both need to be evaluated for
the downstream model.

\subsection{RQ2: Does Symbolic Negative Selection Generalize Across Recommendation Settings?}
\label{sec:rq2-generalization}

We further evaluate our sampling framework on three general recommendation
datasets: KuaiRec~\cite{gao2022kuairec}, MIND~\cite{wu2020mind}, and
RetailRocket~\cite{retailrocket2015}. These datasets provide substantially
different recommendation settings and, importantly, different forms of
behavioral evidence for distinguishing positive and negative interactions.

KuaiRec is a nearly fully observed recommendation dataset collected from
Kuaishou. Almost all 1,411 users are exposed to all 3,327 videos, which makes
it considerably denser than conventional implicit-feedback datasets
\cite{gao2022kuairec}. In our natural KuaiRec formulation, 34.5\% of the
training outcomes are positive. MIND is constructed from Microsoft News
impression logs. Unlike conventional implicit-feedback datasets in which an
unobserved item is simply treated as a potential negative, MIND records which
news articles were actually shown to the user and whether each displayed
article was clicked~\cite{wu2020mind}. Our MIND-small preprocessing contains
50,000 users, 51,282 articles, 17 categories, 156,965 impressions, and a
4.13\% click rate. RetailRocket contains real-world e-commerce events collected
over 4.5 months, including views, add-to-cart events, and transactions. The
original dataset contains 2.76 million events from approximately 1.41 million
visitors~\cite{retailrocket2015}. Because most visitors have extremely short
histories, we use an active-user subset with at least five events, containing
81,620 visitors. Item categories are mapped to their parent categories (268
categories), and purchases constitute 3.10\% of the training interactions.

\begin{table}[t]
\centering
\scriptsize
\setlength{\tabcolsep}{2.6pt}
\caption{
Characteristics of the three public recommendation settings used in RQ2.
The experimental statistics correspond to the processed data used in our
experiments rather than necessarily to the full released dataset.
}
\label{tab:rq3-dataset-stats}
\begin{tabular}{@{}p{1.3cm}p{1.7cm}p{2.3cm}p{1.6cm}@{}}
\toprule
\textbf{Dataset}
& \textbf{Experimental scale}
& \textbf{Observed behavioral signal}
& \textbf{Positive rate} \\
\midrule

KuaiRec
& 1,411 users; 3,327 videos
& Nearly fully observed user--video feedback
& 34.5\% natural \\

MIND
& 50k users; 51,282 articles; 156,965 impressions
& Impression $\rightarrow$ click / non-click
& 4.13\% clicks \\

RetailRocket
& 81,620 active visitors
& View $\rightarrow$ cart $\rightarrow$ transaction
& 3.10\% purchases \\

\bottomrule
\end{tabular}
\end{table}

These settings are useful because the motivation of our symbolic framework is
not simply that recommendation data are imbalanced. The more important problem
is that an unobserved interaction may not be a reliable negative. In the
industrial setting, interactions are sparse and negative preference is not
directly observed. The public datasets in
Table~\ref{tab:rq3-dataset-stats} expose different amounts and types of
behavioral information. KuaiRec provides dense observations, MIND provides
actual impression-level non-clicks, and RetailRocket provides a sequence of
positive behavioral events such as views, carts, and purchases. We therefore
use these datasets to study both the direction of the symbolic evidence and
where symbolic candidate selection is useful.

For downstream evaluation, we report test PR-AUC and Top-1 accuracy.
PR-AUC is particularly appropriate in these settings because positive
interactions are strongly imbalanced in MIND and RetailRocket. Top-1 measures
whether the highest-ranked item in the evaluation candidate set is a positive
target. The two metrics therefore capture complementary behavior: PR-AUC
evaluates ranking across the complete score distribution, whereas Top-1
evaluates the correctness of the model's highest-ranked recommendation.
Results are averaged over seeds 42, 1, and 2.

\paragraph{Negative versus positive symbolic evidence.}

Our original symbolic sampler is \emph{negative-oriented}: rule discovery
identifies user--item regions for which the observed evidence supports treating
an interaction as a defensible negative, and the sampler preferentially draws
from these regions. This formulation follows the motivating industrial
setting, where the primary problem is deciding whether an unobserved
customer--product interaction can safely be added as negative supervision.

The public datasets, however, often contain a different type of information.
For example, MIND explicitly records clicks, while RetailRocket records a
behavioral progression from views to carts and transactions. In such cases it
may be easier to identify evidence that an interaction is \emph{potentially
positive} than to infer that it is definitely negative.

We therefore construct a \textbf{Positive Symbolic} formulation. Instead of
discovering rules associated with negative outcomes, we discover rules
associated with positive outcomes using the same evidence-based rule discovery
principle. These rules do not create additional positive training examples.
Rather, they act as a protection mechanism: if a candidate lies in a
symbolically supported positive region, it is prevented from being selected
as a negative whenever safer candidates are available.
The reported Positive Symbolic condition uses hard tier ordering.

If $E^{-}(u,i)$ denotes negative symbolic evidence and $E^{+}(u,i)$ denotes
positive symbolic evidence, the two formulations differ conceptually as

\begin{equation}
E^{-}(u,i) \uparrow
\quad\Rightarrow\quad
P(i\ \textrm{selected as negative}\mid u)\uparrow,
\end{equation}

whereas Positive Symbolic implements

\begin{equation}
E^{+}(u,i) \uparrow
\quad\Rightarrow\quad
P(i\ \textrm{selected as negative}\mid u)\downarrow.
\end{equation}

Thus, the symbolic framework is unchanged in its purpose---using structured
behavioral evidence to determine negative-candidate validity---but the
direction of the evidence is adapted to the information available in the
dataset.

\begin{table}[t]
\centering
\scriptsize
\setlength{\tabcolsep}{2.4pt}
\caption{
Negative- and positive-oriented symbolic sampling on the natural public
datasets. Best Epoch is the validation-selected checkpoint. Values are mean
$\pm$ sample SD over seeds 42, 1, and 2. Higher PR-AUC and Top-1 are better.
}
\label{tab:rq3-pos-neg}
\begin{tabular}{@{}llrrr@{}}
\toprule
\textbf{Dataset} & \textbf{Sampler}
& \textbf{Test PR-AUC}
& \textbf{Top-1}
& \textbf{Best Epoch} \\
\midrule

KuaiRec
& Random & $0.4617\pm0.0208$ & $0.6249\pm0.0462$ & 4.3 \\
& BNS & $0.5086\pm0.0129$ & $0.6893\pm0.0197$ & 5.0 \\
& DNS & $0.4135\pm0.0065$ & $0.6116\pm0.0356$ & 8.0 \\
& Negative Symbolic & $0.4100\pm0.0057$ & $0.6485\pm0.0186$ & 1.0 \\
& \textbf{Positive Symbolic} & $\mathbf{0.5417\pm0.0010}$ & $\mathbf{0.7425\pm0.0548}$ & \textbf{1.0} \\

\midrule

MIND
& Random & $0.0378\pm0.0005$ & $0.0945\pm0.0025$ & 1.7 \\
& BNS & $0.0385\pm0.0010$ & $0.1014\pm0.0073$ & 1.7 \\
& DNS & $0.0378\pm0.0001$ & $\mathbf{0.1021\pm0.0015}$ & 3.7 \\
& Negative Symbolic & $0.0371\pm0.0002$ & $0.0966\pm0.0027$ & 2.3 \\
& \textbf{Positive Symbolic} & $\mathbf{0.0398\pm0.0011}$ & $0.0993\pm0.0041$ & 5.0 \\

\midrule

RetailRocket
& Random & $0.0502\pm0.0002$ & $0.5495\pm0.0178$ & 3.0 \\
& BNS & $0.0510\pm0.0013$ & $0.5583\pm0.0071$ & 3.3 \\
& \textbf{DNS} & $\mathbf{0.0526\pm0.0014}$ & $\mathbf{0.5600\pm0.0063}$ & 7.3 \\
& Negative Symbolic & $0.0502\pm0.0018$ & $0.5492\pm0.0081$ & 3.3 \\
& Positive Symbolic & $0.0413\pm0.0012$ & $0.5495\pm0.0086$ & 3.3 \\

\bottomrule
\end{tabular}
\end{table}

\paragraph{Results.}
As shown in Table~\ref{tab:rq3-pos-neg}, the direction in which symbolic
evidence is used changes the downstream result. On natural KuaiRec, Positive
Symbolic obtains a test PR-AUC of 0.5417 compared with 0.4100 for the original
negative-oriented sampler. It also exceeds BNS (0.5086), DNS (0.4135), and
Random (0.4617). Top-1 follows the same pattern: Positive Symbolic reaches
0.7425 compared with 0.6485 for Negative Symbolic and 0.6893 for BNS.

An additional difference appears in convergence. On KuaiRec, both symbolic
formulations select their best validation checkpoint after the first epoch,
whereas Random and BNS select checkpoints around epochs 4--5 and DNS at epoch
8. This is consistent with the symbolic candidate pool making the early
training signal easier to separate: the model is not required to learn only
from unrestricted unobserved candidates, but receives negatives already
partitioned using behavioral evidence. We interpret this as faster
\emph{convergence in epochs}; because we do not report full wall-clock
measurements here, we do not claim a corresponding wall-clock speedup.

The MIND result is smaller but follows the same PR-AUC direction. Positive
Symbolic has mean PR-AUC 0.0398, compared with 0.0371 for Negative Symbolic,
0.0378 for Random and DNS, and 0.0385 for BNS. These differences are comparable
to the reported seed variation, so we do not interpret them as a clear
separation between methods. The smaller differences are consistent with the structure of MIND:
non-clicked items are not arbitrary unobserved catalogue items, but items
actually shown in an impression. The MIND format therefore provides a more
direct negative signal than a conventional implicit-feedback matrix. Among
sampled candidates with an observed label, all major samplers already achieve
approximately 98.5--99.2\% negative precision, leaving substantially less room
for a candidate-validity mechanism to improve the downstream model.

RetailRocket gives the opposite result. DNS is strongest, reaching 0.0526
PR-AUC and 0.5600 Top-1, while Positive Symbolic reaches only 0.0413 PR-AUC.
This behavior is explained more directly by the discovered positive rules.
Table~\ref{tab:rq3-positive-quality} reports whether the region protected by
the strongest positive rules is actually enriched for positives.

\begin{table}[t]
\centering
\scriptsize
\setlength{\tabcolsep}{2.5pt}
\caption{
Quality of the strongest positive-symbolic protection region. Positive
Precision is the observed positive rate within the strongly protected region;
Base Rate is the positive rate over the labelled candidate pool. Lift above
1 indicates that the rule identifies a region enriched for positives.
Positive precision and recall are identical across seeds 42, 1, and 2
(sample SD 0).
}
\label{tab:rq3-positive-quality}
\begin{tabular}{@{}lrrrr@{}}
\toprule
\textbf{Dataset}
& \textbf{Positive Prec.}
& \textbf{Base Rate}
& \textbf{Lift}
& \textbf{Positive Recall} \\
\midrule

KuaiRec natural
& 0.1438
& 0.1327
& 1.08$\times$
& 0.0397 \\

MIND
& 0.0181
& 0.0147
& 1.23$\times$
& 0.3808 \\

RetailRocket
& 0.0052
& 0.0105
& 0.49$\times$
& 0.0802 \\

\bottomrule
\end{tabular}
\end{table}

The protected MIND region has a 1.23$\times$ positive-rate lift over the base
candidate population, which is consistent with the improvement of Positive
Symbolic in Table~\ref{tab:rq3-pos-neg}. On RetailRocket, however, the
strongly protected region has only 0.49$\times$ the base positive rate.
Thus, in the representation available to our rule-discovery procedure,
the positive rules do not identify a positively enriched subset of
RetailRocket candidates. Protecting those candidates consequently removes
useful negatives rather than reliably protecting likely purchases.

This result is also consistent with the structure of RetailRocket. The
original dataset contains approximately 1.41 million visitors, but the median
visitor has only one event. Even after restricting the experiment to users
with at least five events, individual histories remain short relative to the
large item catalogue. The discovered user-conditional symbolic rules
therefore have substantially less evidence per user than in KuaiRec. In our
diagnostics, a large fraction of RetailRocket symbolic selections fall back
from the primary symbolic pool. DNS, in contrast, does not need a
well-supported user--category rule. It evaluates a candidate set using the
current model and selects hard negatives. Hard-negative sampling is known to
be particularly connected to Top-$K$ ranking objectives
\cite{zhang2013dns,shi2023hard}, which is consistent with DNS remaining the
strongest natural RetailRocket sampler.

These results show that ``positive'' or ``negative'' symbolic sampling should
not be treated as a fixed choice. The correct direction depends on the
available behavioral evidence. When a dataset supplies reliable evidence for
positive preference, positive symbolic rules can be used to protect likely
positives. When reliable negative regions can be discovered, the original
negative-oriented formulation can instead be used directly.

\paragraph{Symbolic candidate selection as a sampling layer.}

The previous experiment treats Symbolic as a complete sampler: the symbolic
pipeline orders the candidate regions and directly supplies the $K$ negatives
used for each positive training example. We next test a more general use of
the framework.

Our motivation is that symbolic reasoning can instead operate as an
\emph{implicit candidate-selection layer} underneath an arbitrary negative
sampling strategy. We separate two questions:

\begin{enumerate}
    \item \emph{Candidate validity:} which interactions are admissible
    negatives given the available behavioral evidence?
    \item \emph{Candidate preference:} among those admissible candidates,
    which negative should the downstream sampling strategy select?
\end{enumerate}

The symbolic pipeline answers the first question and the existing sampler
answers the second. Let $\mathcal{C}(u)$ be the original candidate set for
user $u$. Symbolic reasoning first constructs

\begin{equation}
\begin{aligned}
\mathcal{C}_{\mathrm{sym}}(u)
= \{i\in\mathcal{C}(u):{}&
i\textrm{ satisfies the symbolic}\\
&\textrm{candidate criterion}\}.
\end{aligned}
\end{equation}

Random + Symbolic then samples uniformly from
$\mathcal{C}_{\mathrm{sym}}(u)$. BNS + Symbolic applies the original Bayesian
weights only inside this candidate set,

\begin{equation}
P(i^{-}=i\mid u)
\propto
\mathbb{I}[i\in\mathcal{C}_{\mathrm{sym}}(u)]
\,w_{\mathrm{BNS}}(i),
\end{equation}

while DNS + Symbolic first restricts the candidate search and then selects a
hard negative,

\begin{equation}
i^{-}
=
\arg\max_{i\in\mathcal{C}_{\mathrm{sym}}(u)}
f_{\theta}(u,i).
\end{equation}

Thus, we do not change how BNS or DNS ranks candidates. The conceptual change
is that they no longer operate over every unobserved candidate. Symbolic
reasoning first decides which interactions are sufficiently supported to
enter the sampler's search space.

\begin{table}[t]
\centering
\scriptsize
\setlength{\tabcolsep}{2.4pt}
\caption{
Symbolic candidate selection used as a layer beneath existing samplers on
the natural public datasets. BNS-S + Symbolic uses the smoothed BNS
implementation, since the unsmoothed symbolic wrapper was not run on all
three natural datasets. Values are mean $\pm$ sample SD over seeds 42, 1, and 2.
}
\label{tab:rq3-wrapper-natural}
\begin{tabular}{@{}llrr@{}}
\toprule
\textbf{Dataset} & \textbf{Sampler}
& \textbf{Test PR-AUC}
& \textbf{Top-1} \\
\midrule

KuaiRec
& Random & $0.4617\pm0.0208$ & $0.6249\pm0.0462$ \\
& Random + Symbolic & $0.3546\pm0.0038$ & $0.6107\pm0.0035$ \\
& BNS-S + Symbolic & $0.3843\pm0.0089$ & $0.5785\pm0.0155$ \\
& DNS + Symbolic & $0.3544\pm0.0010$ & $0.6475\pm0.0741$ \\
& Symbolic Alone & $0.4100\pm0.0057$ & $\mathbf{0.6485\pm0.0186}$ \\

\midrule

MIND
& Random & $0.0378\pm0.0005$ & $0.0945\pm0.0025$ \\
& Random + Symbolic & $0.0380\pm0.0010$ & $0.0959\pm0.0058$ \\
& \textbf{BNS-S + Symbolic} & $\mathbf{0.0384\pm0.0012}$ & $0.0989\pm0.0047$ \\
& DNS + Symbolic & $0.0377\pm0.0004$ & $\mathbf{0.0991\pm0.0066}$ \\
& Symbolic Alone & $0.0371\pm0.0002$ & $0.0966\pm0.0027$ \\

\midrule

RetailRocket
& Random & $0.0502\pm0.0002$ & $0.5495\pm0.0178$ \\
& Random + Symbolic & $0.0502\pm0.0011$ & $0.5500\pm0.0160$ \\
& BNS-S + Symbolic & $0.0517\pm0.0014$ & $\mathbf{0.5619\pm0.0063}$ \\
& \textbf{DNS + Symbolic} & $\mathbf{0.0524\pm0.0007}$ & $0.5520\pm0.0071$ \\
& Symbolic Alone & $0.0502\pm0.0018$ & $0.5492\pm0.0081$ \\

\bottomrule
\end{tabular}
\end{table}

\paragraph{Results.}
Table~\ref{tab:rq3-wrapper-natural} shows that candidate filtering does not
uniformly improve dense natural data, but it demonstrates that the symbolic
pipeline can be composed with existing sampling procedures without changing
their sampling objective.

On KuaiRec natural, filtering before the conventional samplers is harmful:
Random + Symbolic, BNS-S + Symbolic, and DNS + Symbolic all have lower
PR-AUC than their unrestricted counterparts. KuaiRec natural is the densest
condition considered here, with a 34.5\% positive rate and a nearly
fully-observed matrix. In this regime, item-level interaction statistics have
substantial support. BNS therefore already receives a useful marginal signal,
and aggressively reducing the candidate pool can discard informative
negatives. This is consistent with BNS being substantially stronger on the
natural distribution than the original symbolic sampler.

On MIND, the differences between the filtered and unfiltered variants are
small. Random + Symbolic changes PR-AUC from 0.0378 to 0.0380 and BNS-S +
Symbolic obtains 0.0384. This small magnitude is consistent with the
impression-level nature of MIND: because a non-click corresponds to an item
that was actually exposed, the candidate pool is already comparatively clean.

RetailRocket provides a more useful example of compositional behavior.
BNS-S + Symbolic reaches 0.0517 PR-AUC and 0.5619 Top-1, compared with
0.0503 PR-AUC and 0.5550 Top-1 for smoothed BNS without symbolic filtering.
DNS + Symbolic reaches 0.0524 PR-AUC, nearly equal to unrestricted DNS at
0.0526. Thus, symbolic filtering can be placed underneath conventional
sampling without replacing the sampling objective. In particular, BNS still
uses propensity-derived weights, while DNS still selects hard examples; the
symbolic component only changes the set over which those operations are
performed.

\paragraph{Does symbolic filtering actually produce cleaner negatives?}

To evaluate the candidate-selection mechanism independently of downstream
PR-AUC, we additionally measure \emph{negative selection precision}. Because
many catalogue pairs in public recommendation datasets are unlabelled, we
compute this metric only over selected candidates for which a clean or
observed ground-truth outcome is available:

\begin{equation}
\mathrm{Prec}_{-}^{\mathrm{labelled}}
=
\frac{
\#\{\textrm{selected candidates with } y=0\}
}{
\#\{\textrm{selected candidates with known }y\}
}.
\end{equation}

The complementary quantity is false-negative contamination,

\begin{equation}
\mathrm{Contam}^{\mathrm{labelled}}
=
1-\mathrm{Prec}_{-}^{\mathrm{labelled}}.
\end{equation}

This avoids counting an unlabelled item as an incorrect negative merely
because its true outcome is unknown.

The effect is clearest under incomplete positive feedback, where clean labels
are available for the artificially hidden positives.

\begin{table}[t]
\centering
\scriptsize
\setlength{\tabcolsep}{2.3pt}
\caption{
Quality of selected negatives on KuaiRec MNAR-75. Negative Precision is
computed among selected candidates with an available clean label. False-Neg.
is its complementary positive-contamination rate. Higher precision and lower
false-negative rate are better. Values are mean $\pm$ sample SD in percentage
points over seeds 42, 1, and 2.
}
\label{tab:rq3-neg-precision}
\begin{tabular}{@{}lrr@{}}
\toprule
\textbf{Sampler}
& \textbf{Negative Prec. $\uparrow$}
& \textbf{False-Neg. $\downarrow$} \\
\midrule

Random
& $95.26\pm0.10\%$ & $4.74\pm0.10\%$ \\

BNS
& $95.55\pm0.08\%$ & $4.45\pm0.08\%$ \\

DNS
& $93.10\pm0.02\%$ & $6.90\pm0.02\%$ \\

Negative Symbolic
& $96.84\pm0.14\%$ & $3.16\pm0.14\%$ \\

Positive Symbolic
& $\mathbf{99.70\pm0.17\%}$ & $\mathbf{0.30\pm0.17\%}$ \\

Random + Symbolic
& $96.43\pm0.12\%$ & $3.57\pm0.12\%$ \\

BNS + Symbolic
& $96.45\pm0.21\%$ & $3.55\pm0.21\%$ \\

DNS + Symbolic
& $94.78\pm0.10\%$ & $5.22\pm0.10\%$ \\

\bottomrule
\end{tabular}
\end{table}

Table~\ref{tab:rq3-neg-precision} shows that symbolic candidate filtering
changes the quality of the negatives supplied to the model. Under MNAR-75,
BNS selects true negatives with 95.55\% precision; adding symbolic candidate
selection increases this to 96.45\%. DNS has the lowest negative precision
among the three conventional samplers, 93.10\%, which is consistent with the
known tradeoff in hard-negative sampling: selecting examples that the current
model finds difficult can also move the sampler closer to unobserved
positives. After symbolic filtering, DNS precision increases to 94.78\%.

The DNS analysis makes this distinction clearer. DNS and DNS + Symbolic both
select approximately the hardest candidates within their own eligible pools;
the symbolic layer therefore does not weaken the DNS rule itself. Instead, it
changes the pool in which hardness is searched. Under MNAR-75, overall
positive contamination falls from 0.00832 for DNS to 0.00441 for DNS +
Symbolic. Among labelled candidates it falls from 6.90\% to 5.22\%. Even
within the hardest 95--100\% candidate band, labelled contamination falls
from 9.88\% to 7.37\%. Thus, the symbolic layer can remove part of the
false-negative risk before DNS performs its usual hardness selection.

Positive Symbolic produces the cleanest negative set in this experiment:
99.70\% of its labelled selected negatives are true negatives. This is
consistent with its construction. Instead of asking which candidates have
negative evidence, it begins by excluding regions with positive symbolic
evidence and samples from regions in which such evidence is absent.

\paragraph{Robustness to sparse and incomplete positive feedback.}

Finally, we directly test the setting that motivates our method: recommendation
data in which positive feedback is either rare or incompletely observed.
We distinguish two different forms of difficulty rather than referring to both
simply as ``skew.''

The \textbf{clean sparsity} experiment changes the prevalence of positive
examples without changing their labels. We construct nested KuaiRec conditions
with approximately 5\%, 2\%, and 1\% positive outcomes. A 5\% condition
therefore contains substantially more observed positive evidence per item than
a 1\% condition, but every retained label remains correct.
The construction also applies category-volume thinning with exponent 1.5 and
per-category positive skew, so the cells differ in category volume as well as
overall positive prevalence.

The second experiment introduces missing-positive noise.
\textbf{MNAR-75} denotes \emph{Missing Not At Random at 75\%}. Starting from
the KuaiRec 5\% condition, we hide 75\% of the positive interactions in the
training data. Missingness is mildly tilted toward high-volume items,

\begin{equation}
P(\textrm{hide }(u,i))
\propto
\left(
\frac{\mathrm{volume}(i)}
{\max_j \mathrm{volume}(j)}
\right)^{0.25},
\end{equation}

subject to the experiment's probability cap. Hidden positives are made
indistinguishable from observed negatives on the features available to the
sampling pipeline. Rules are rediscovered from the corrupted training data;
the clean label is retained only in a sidecar for post-hoc evaluation and is
never available to the sampler. Validation and test distributions remain
unchanged.

The 1\% experiment therefore asks whether the sampler remains useful when
positive evidence is \emph{sparse}. MNAR-75 asks a different question:
whether it remains useful when the apparent negative pool itself contains a
substantial number of \emph{missing positives}.

\begin{table}[t]
\centering
\scriptsize
\setlength{\tabcolsep}{2.5pt}
\caption{
Downstream performance as KuaiRec positive feedback becomes sparse or
incomplete. The 5\% and 1\% settings contain clean labels. MNAR-75 starts
from the 5\% condition and hides 75\% of positive training interactions.
Values are mean $\pm$ sample SD over seeds 42, 1, and 2.
}
\label{tab:rq3-sparse}
\begin{tabular}{@{}llrr@{}}
\toprule
\textbf{Setting} & \textbf{Sampler}
& \textbf{Test PR-AUC}
& \textbf{Top-1} \\
\midrule

5\% clean
& Random & $\mathbf{0.1531\pm0.0030}$ & $0.3598\pm0.0067$ \\
& BNS & $0.1357\pm0.0475$ & $0.3596\pm0.0644$ \\
& DNS & $0.0617\pm0.0005$ & $0.2353\pm0.0161$ \\
& Symbolic & $0.1085\pm0.0034$ & $0.2979\pm0.0139$ \\

\midrule

1\% clean
& Random & $0.0139\pm0.0007$ & $0.0340\pm0.0046$ \\
& BNS & $0.0150\pm0.0003$ & $0.0373\pm0.0043$ \\
& DNS & $0.0131\pm0.0008$ & $0.0553\pm0.0097$ \\
& \textbf{Symbolic} & $\mathbf{0.0213\pm0.0004}$ & $\mathbf{0.0768\pm0.0145}$ \\

\midrule

5\% + MNAR-75
& Random & $0.0940\pm0.0051$ & $0.2400\pm0.0120$ \\
& BNS & $0.1075\pm0.0010$ & $0.2773\pm0.0082$ \\
& DNS & $0.0799\pm0.0006$ & $0.3312\pm0.0143$ \\
& \textbf{Symbolic} & $\mathbf{0.1337\pm0.0006}$ & $\mathbf{0.4115\pm0.0126}$ \\

\bottomrule
\end{tabular}
\end{table}

\paragraph{Results.}
Table~\ref{tab:rq3-sparse} shows a clear regime change. At 5\% clean
positives, Symbolic is not the strongest sampler. Random reaches 0.1531
PR-AUC, BNS reaches 0.1357, and Symbolic reaches 0.1085. This is a useful
boundary condition rather than a contradictory result. When sufficient clean
positive observations remain, item- and model-based samplers have enough
information to construct useful negatives, and the additional symbolic
restriction can remove informative candidates.

When the clean positive rate is reduced from 5\% to 1\%, this ordering
changes. Symbolic obtains 0.0213 PR-AUC compared with 0.0150 for BNS,
0.0139 for Random, and 0.0131 for DNS. It also has the highest Top-1 score,
0.0768. In this regime, the marginal item-level statistics used by BNS have
very little positive support. As the positive count approaches zero for many
items, their estimated negative weights become increasingly similar, and BNS
approaches uniform sampling. Symbolic selection instead aggregates evidence
over structured regions of the user--item space and can abstain from regions
without sufficient support. The gain at 1\% is therefore consistent with the
method's original motivation: structured evidence becomes more valuable when
fine-grained observed counts become sparse.

The MNAR-75 result is stronger because it adds a second source of difficulty.
Here the sampler does not merely observe fewer positives; many actual positives
are placed into the apparent negative pool. Symbolic reaches 0.1337 PR-AUC
and 0.4115 Top-1, compared with 0.1075 and 0.2773 for BNS. Random reaches
0.0940 PR-AUC, while DNS reaches 0.0799.

This difference is also visible before downstream training.
Table~\ref{tab:rq3-neg-precision} shows that BNS selects 4.45\% false
negatives among labelled candidates under MNAR-75, whereas Negative Symbolic
reduces this to 3.16\%. Positive Symbolic reduces it further to 0.30\%.
Thus, when positives are hidden, the symbolic pipeline is able to remove
candidate regions in which the apparent negative label is contradicted by
other evidence.

The BNS result is particularly informative. BNS is designed to distinguish
true from false negatives using an estimated negative signal
\cite{liu2023bns}. When item-level positive counts are well supported, as in
dense KuaiRec, this signal can be effective. Under sparse or MNAR feedback,
however, hiding a positive also lowers the item's observed positive rate.
The item can therefore appear \emph{more negative} precisely because its
positive evidence has gone missing. The structured symbolic pipeline is less
dependent on this single marginal count and is therefore more robust in the
MNAR condition.

DNS exhibits a different failure mode. DNS deliberately selects high-scoring
hard negatives, which can improve Top-$K$ optimization
\cite{zhang2013dns,shi2023hard}. When all unobserved candidates are reliable,
that hardness is useful. When the pool contains hidden positives, however,
the hardest apparent negatives are also natural places for false negatives to
occur. This is visible directly in our diagnostics: DNS has 6.90\% labelled
false-negative contamination under MNAR-75, the highest of the conventional
samplers in Table~\ref{tab:rq3-neg-precision}. Symbolic candidate filtering
reduces it to 5.22\% while leaving DNS responsible for the final hardness
selection.

\paragraph{Candidate filtering in the sparse regime.}

The same stress experiment also clarifies why the candidate-selection
formulation is useful even when it is not beneficial on dense natural
KuaiRec. At 1\% clean positives, applying the symbolic candidate layer raises
Random from $0.0139\pm0.0007$ to $0.0191\pm0.0001$ PR-AUC, smoothed BNS from
$0.0142\pm0.0008$ to $0.0194\pm0.0003$, and DNS from $0.0131\pm0.0008$ to
$0.0194\pm0.0001$. The original Symbolic sampler obtains $0.0213\pm0.0004$.

Under MNAR-75, Random increases from $0.0940\pm0.0051$ to
$0.1277\pm0.0003$ after symbolic candidate filtering; smoothed BNS increases
from $0.1017\pm0.0023$ to $0.1279\pm0.0005$; and DNS increases from
$0.0799\pm0.0006$ to $0.1247\pm0.0011$. Negative Symbolic itself reaches
$0.1337\pm0.0006$. Therefore, the wrapper experiment does not
suggest that every sampler should always be symbolically filtered. Rather, it
shows that the symbolic layer becomes valuable precisely when candidate
validity is uncertain.

This distinction is central to the proposed framework. Conventional negative
sampling methods primarily answer \emph{which negative should be useful for
learning?} Random emphasizes coverage, BNS uses estimated negative
probability, and DNS emphasizes hardness. Our symbolic stage answers an
earlier question: \emph{which interactions are sufficiently supported to be
treated as negatives in the first place?} The two decisions are therefore
complementary rather than competing.

\begin{figure}[t]
    \centering
    \includegraphics[width=\columnwidth]{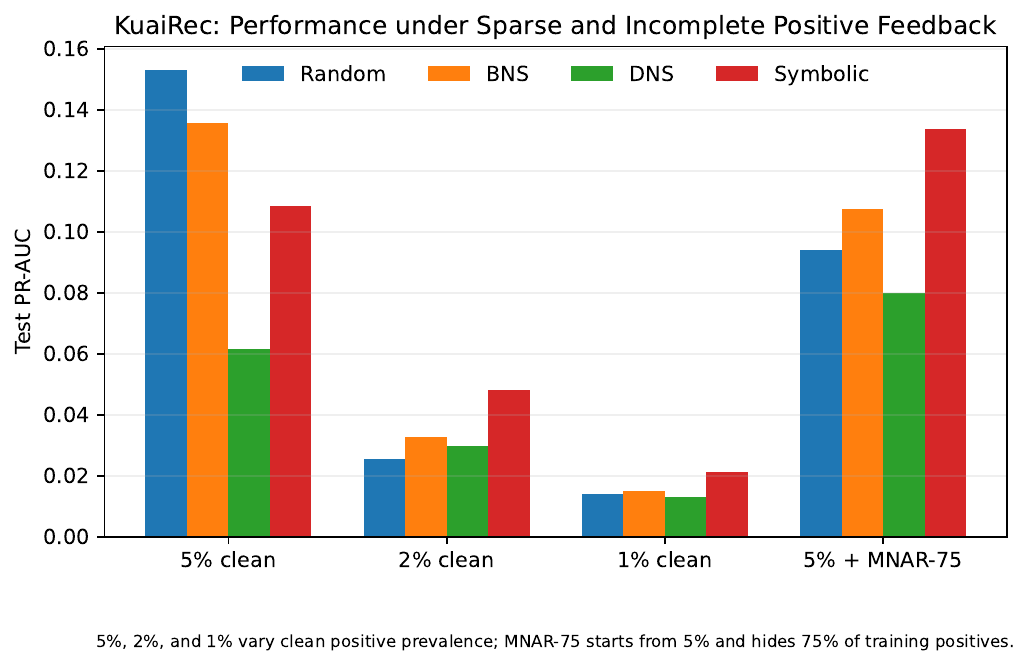}
    \caption{
    Test PR-AUC on KuaiRec as positive feedback becomes sparse or incomplete.
    The first three groups (5\%, 2\%, and 1\%) vary clean positive prevalence.
    The fourth group, 5\% + MNAR-75, instead starts from the 5\% condition and
    hides 75\% of positive training interactions. The MNAR point should
    therefore be interpreted as a missing-feedback stress condition rather than
    as another point on the positive-rate axis.
    }
    \Description{A grouped bar chart comparing test PR-AUC for Random, BNS,
    DNS, and Symbolic sampling on KuaiRec with 5 percent, 2 percent, and
    1 percent clean positive feedback and with 75 percent of positive training
    interactions hidden under MNAR.}
    \label{fig:rq3-sparse-mnar}
\end{figure}

Figure~\ref{fig:rq3-sparse-mnar} summarizes this transition. On clean
KuaiRec, Symbolic becomes more competitive as the positive rate decreases:
at 5\% it trails Random and BNS, at 2\% it reaches $0.0482\pm0.0006$ PR-AUC
compared with $0.0327\pm0.0036$ for BNS and $0.0255\pm0.0003$ for Random,
and at 1\% it reaches 0.0213
compared with 0.0150 and 0.0139, respectively. The MNAR-75 condition produces
the same qualitative conclusion through a different mechanism: rather than
reducing the number of positive examples directly, it makes the observed
negative pool less trustworthy.

\paragraph{Summary.}

Taken together, RQ2 shows that symbolic negative selection should be
understood as a mechanism for incorporating structured behavioral evidence
into candidate construction rather than as a universally dominant sampling
distribution.

First, the appropriate direction of the rules depends on the type of evidence
available in the data. Negative-oriented rules are natural when we can
identify supported regions of negative behavior, while positive-oriented
rules can instead protect candidates when clicks, purchases, or other
positive behaviors provide the stronger signal. On KuaiRec and MIND, this
positive-oriented formulation improves over the original negative symbolic
sampler. RetailRocket provides the corresponding boundary condition: when
the discovered positive rules are not enriched for actual positives, positive
protection does not help.

Second, symbolic reasoning can be separated from the downstream sampling
strategy. Random, BNS, and DNS can all operate within a symbolically admitted
candidate pool without modifying their underlying sampling objective. This
provides a general way to use symbolic reasoning as a candidate-selection
layer rather than as a replacement for existing negative samplers.

Third, the benefit depends on the reliability of the apparent negative pool.
When KuaiRec is dense and clean, conventional samplers are competitive or
stronger. When positive evidence becomes sparse at 1\%, or when 75\% of
positive training feedback becomes missing under MNAR, the symbolic sampler
has higher PR-AUC and Top-1 than Random, BNS, and DNS. Candidate-quality
diagnostics show the corresponding
change in the supervision supplied to the recommender: under MNAR-75,
Negative Symbolic raises labelled negative precision from 95.55\% for BNS to
96.84\%, while Positive Symbolic reaches 99.70\%.

Overall, these results support the motivating use of symbolic evidence for
recommendation settings in which missing or sparse observations make
unobserved interactions unreliable negatives. They also identify the
boundary of the approach: when negatives are directly observed or the
interaction matrix is dense enough that conventional estimates are reliable,
symbolic filtering need not improve downstream accuracy.

\section{Conclusion}
\label{sec:conclusion}

We study how observed customer behavior can support the selection of negative
examples for recommendation. Our method expresses the behavior as symbolic
rules and evaluates each rule using observed support, purchase-rate
uncertainty, information beyond simpler conditions, and differences across
products. After these statistical checks, the LLM relates each rule to the
business objective and domain knowledge supplied by data scientists. The
retained rules are ranked by their evidence scores and recorded in a
domain-knowledge report so that data scientists can examine the patterns used
for selection.

The rules identify higher-precision negative candidates than matched random
selection in the industrial setting and three public candidate-quality
datasets. Across five public recommendation datasets overall, our experiments
evaluate candidate quality and characterize when symbolic candidate selection
improves downstream recommendation. Their use in
training improves downstream recommendation at several matched budgets, but
adding more negatives does not consistently help. The business review also
reduces candidate precision in the reported ablation. Candidate quality,
business interpretation, and downstream performance therefore need separate
evaluation. The method makes the evidence behind negative selection available
for both training and review.

\section{Ethical Considerations}
\label{sec:ethics}

The method should not be interpreted as recovering a customer's unobserved dislike
of a product. It identifies observationally supported candidates for downstream
training, and its outputs can reflect historical exposure, eligibility,
measurement, and deployment biases. Any use of customer data must follow the
applicable privacy, security, and retention requirements. In deployment, the
rule reports and model outputs should support human review rather than determine
eligibility, access, pricing, or other high-impact decisions. The rule-level
evidence and the separate domain review are intended to make potential bias and
spurious correlations easier to inspect. The language model is limited to a
post hoc domain audit and must not receive unnecessary customer-identifying
information or override the statistical admission criteria.
\bibliographystyle{ACM-Reference-Format}

\clearpage
\appendix
\section{Additional Candidate-Quality Evaluation}
\label{app:additional-datasets}

\subsection{Industrial Candidate Quality}
\label{app:candidate-quality}

The downstream experiments measure whether different negative-sampling
strategies improve recommendation performance. Here, we ask a complementary
question: \emph{does the symbolic procedure identify higher-quality negative
candidates in the first place?} This distinction matters because an
unobserved interaction is not necessarily evidence of disinterest; it may
also reflect timing, exposure, or missing context. We therefore evaluate
candidate quality independently of downstream model performance.

\paragraph{Held-out evaluation.}
We evaluate on held-out customer contacts for which the product was offered
and a 30-day purchase outcome is observed. The evaluation pool contains 5,074
examples: 4,035 non-purchases and 1,039 purchases, corresponding to a
79.5\% non-purchase rate.

We use the observed 30-day outcome to check how many selected negatives
correspond to purchases. This evaluates selection on observed outcomes rather
than assuming that unobserved interactions are true negatives. For a selected candidate
set $\mathcal{N}$, we define negative-candidate precision as

\begin{equation}
\mathrm{Prec}_{-}(\mathcal{N})
=
\frac{
    \sum_{(u,p)\in\mathcal{N}}
    \mathbb{I}[y_{u,p}=0]
}{
    |\mathcal{N}|
},
\end{equation}

where $y_{u,p}=1$ denotes an observed purchase and $y_{u,p}=0$ denotes
no purchase within the observation period. We additionally report the
number of purchases selected into the candidate set,

\begin{equation}
C(\mathcal{N})
=
\sum_{(u,p)\in\mathcal{N}}
\mathbb{I}[y_{u,p}=1],
\end{equation}

and candidate coverage,

\begin{equation}
\mathrm{Coverage}(\mathcal{N})
=
\frac{|\mathcal{N}|}{|\mathcal{E}|},
\end{equation}

where $\mathcal{E}$ is the set of eligible evaluation pairs. Higher negative
precision and lower contamination indicate cleaner candidate sets, while
coverage measures how broadly the method can identify candidates. We do not
interpret non-purchase as definitive user dislike; rather, these metrics
measure how often a candidate-selection procedure avoids contradicting an
observed purchase.

\paragraph{Selection controls.}
Higher precision could arise simply from selecting fewer pairs or from
concentrating on products with naturally high non-purchase rates. We
therefore compare each symbolic stage against baselines with the same
candidate budget.

Uniform random selects the same number of pairs from the eligible pool.
Popularity-based sampling allocates the same budget according to product-level
purchase frequency. The \emph{product-matched random} control
preserves the exact number of selected pairs for every product and randomizes
only which customers are selected within each product. Thus, differences
against this baseline isolate within-product candidate selection rather than
changes in candidate budget or product composition.

\begin{table*}[t]
\centering
\small
\caption{
Candidate quality under successive evidence constraints.
All baselines use the same candidate-pair budget as the corresponding
symbolic stage; product-matched random additionally preserves the exact
per-product allocation. Each stage enforces a distinct validity criterion and
does not necessarily optimize negative-candidate precision monotonically. Higher
negative-candidate precision and fewer selected purchases are better.
}
\label{tab:candidate_quality}
\resizebox{\textwidth}{!}{
\begin{tabular}{lcccc}
\toprule
\textbf{Metric}
& \textbf{Scope + Evidence}
& \textbf{+ Hierarchy}
& \textbf{+ Product Relevance}
& \textbf{+ Domain Review} \\
\midrule

Rules retained
& 111 & 67 & 55 & 28 \\

Candidate pairs
& 2,034 & 1,637 & 1,098 & 255 \\

Coverage
& 40.09\% & 32.26\% & 21.64\% & 5.03\% \\

\midrule
\textbf{Ours: Candidate Precision $\uparrow$}
& \textbf{0.9223}
& \textbf{0.9163}
& \textbf{0.9490}
& \textbf{0.8902} \\

\textbf{Ours: Purchases Selected $\downarrow$}
& \textbf{158}
& \textbf{137}
& \textbf{56}
& \textbf{28} \\

\midrule
Uniform Random: Precision
& 0.7950 & 0.7947 & 0.7949 & 0.7932 \\

Uniform Random: Expected Purchases
& 417 & 336 & 225 & 53 \\

Popularity: Precision
& 0.6988 & 0.6985 & 0.6984 & 0.6989 \\

Popularity: Expected Purchases
& 613 & 494 & 331 & 77 \\

Product-Matched Random: Precision
& 0.8578 & 0.8349 & 0.9136 & 0.8154 \\

Product-Matched Random: Expected Purchases
& 289.2 & 270.2 & 94.9 & 47.1 \\

\bottomrule
\end{tabular}
}
\end{table*}

\paragraph{Results.}
Scope and Evidence Strength select 2,034 candidates with precision 0.9223,
compared with 0.7950 for uniform random and 0.6988 for popularity-based
selection (Table~\ref{tab:candidate_quality}). Product-matched random gives
0.8578 precision. The relative gain over this control is 7.5\%, with the
candidate count and product allocation held fixed.

Adding Hierarchical Contribution reduces the candidate count to 1,637 and
precision to 0.9163. Adding Product Relevance gives the highest precision,
0.9490, while retaining 1,098 candidates. These candidates include 56 observed
purchases, compared with 94.9 expected under product-matched random selection.
Thus, the product-specific rules select fewer purchases within the same
product allocation. The hierarchy check removes rules without enough
additional information, but does not increase precision in this comparison.

Domain Review further reduces the set to 255 candidates and precision
to 0.8902. This remains above the matched control's 0.8154, but is lower than
precision before the LLM review. The review filters rules according to their
proposed business meaning; the reported results do not show that it improves
candidate precision. The accuracy of those business interpretations is not
measured by the purchase-outcome evaluation.

\subsection{Generalization Across Public Recommendation Datasets}
\label{app:public-generalization}

We next evaluate the method beyond the B2B setting using three public
datasets. KuaiRec provides densely observed user--item interactions
\cite{gao2022kuairec}; KuaiRand provides randomized exposure
\cite{gao2022kuairand}; and Santander Product Recommendation has a large
customer population, a small product catalog, and rich customer attributes
\cite{santander2016product}.

Across all datasets, we compare \textbf{Symbolic} selection against uniform
random, popularity-based, matched-random, and Bayesian negative selection.
Matched-random preserves the candidate budget and the product or category
composition of the symbolic candidate set, while randomizing which users are
selected. Candidate budgets are 242,896 for KuaiRec, 232,804 for KuaiRand,
and 3,706,652 for Santander, and are matched across methods within each
dataset. We report negative-candidate precision, defined as the proportion of selected
candidates with a negative held-out outcome, and contamination, which
counts selected candidates with a positive held-out outcome.
\begin{table}[t]
\centering
\scriptsize
\setlength{\tabcolsep}{2pt}
\caption{
Negative candidate quality across public recommendation datasets.
Candidate budgets are stated in the text. Cat.-Matched and Prod.-Matched
preserve category and product composition, respectively. Higher precision and
lower contamination are better. Symbolic and Bayesian results use seed 42;
no across-seed SD is available for this table.
}
\label{tab:rq3_generalization}
\begin{tabular}{@{}llrrr@{}}
\toprule
\textbf{Dataset}
& \textbf{Method}
& \textbf{Rules}
& \textbf{Precision $\uparrow$}
& \textbf{Contam. $\downarrow$} \\
\midrule

KuaiRec
& Uniform & -- & 0.6771 & 78,441 \\
KuaiRec
& Popularity & -- & 0.6761 & 78,684 \\
KuaiRec
& Cat.-Matched & -- & 0.6868 & 76,075 \\
KuaiRec
& Bayesian & -- & 0.7611 & 58,026 \\
KuaiRec
& \textbf{Symbolic} & \textbf{43} & \textbf{0.7722} & \textbf{55,320} \\

\midrule

KuaiRand
& Uniform & -- & 0.8238 & 41,023 \\
KuaiRand
& Popularity & -- & 0.7803 & 51,158 \\
KuaiRand
& Cat.-Matched & -- & 0.8329 & 38,909 \\
KuaiRand
& Bayesian & -- & 0.8395 & 37,365 \\
KuaiRand
& \textbf{Symbolic} & \textbf{95} & \textbf{0.9030} & \textbf{22,585} \\

\midrule

Santander
& Uniform & -- & 0.9983 & 6,402 \\
Santander
& Popularity & -- & 0.9940 & 22,331 \\
Santander
& Prod.-Matched & -- & 0.9986 & 5,136 \\
Santander
& Bayesian & -- & 0.9982 & 6,625 \\
Santander
& \textbf{Symbolic} & \textbf{396} & \textbf{0.9992} & \textbf{2,917} \\

\bottomrule
\end{tabular}
\end{table}

\paragraph{Results.}
Table~\ref{tab:rq3_generalization} shows consistent gains across the three
settings. On KuaiRec, Symbolic reaches 0.7722 negative-candidate precision, exceeding
uniform random (0.6771), category-matched random (0.6868), and Bayesian
selection (0.7611). On KuaiRand, it reaches 0.9030 and reduces contamination
from 38,909 under category-matched random to 22,585. This comparison holds the candidate budget and category allocation fixed.

Santander has a much higher underlying negative rate, so absolute precision
differences are small. Symbolic still achieves the highest precision, 0.9992,
and admits 2,917 positive acquisitions, a 43.2\% reduction from 5,136 under
product-matched random and below 6,625 under Bayesian selection. Across interaction densities, exposure
mechanisms, and catalog structures, the matched-random controls indicate that
the gains are not explained only by budget or product composition.

\subsection{LLM Domain Review}
The public review code uses \texttt{gpt-4o-mini} with temperature 0 and JSON
output. It makes one call for each deduplicated rule--product pair that passes
all four statistical criteria. The prompt contains a plain-language rule, the
target category, the claim being reviewed, and feature definitions. A
banned-token guard blocks requests containing stage names, scores, counts,
rates, or evaluation results. The reported public experiments do not use this
gate. The released public-data code implements a simplified variant with three
labels (relevant, not relevant, unresolved); no reported public result depends
on this review. The industrial review used GPT-5.4 with greedy decoding for
reasoning. It used one call per rule, followed by one full-review call to
double-check the complete set.

Figure~\ref{fig:llm-prompt} shows the prompt template. The system message
states the reviewer role, the domain knowledge, and the labeling criteria. The
user message is one structured record per rule--product pair. Bracketed fields
are filled from the rule and from the trusted domain knowledge. No field
contains a score, count, rate, or evaluation result.

\begin{figure*}[t]
\centering
\footnotesize
\begin{minipage}{0.95\textwidth}
\textbf{System message}
\begin{verbatim}
You are a domain analyst for <PRODUCT PORTFOLIO DESCRIPTION>.

Domain knowledge:
<TRUSTED DOMAIN KNOWLEDGE: what each product does, who it is intended for, and known usage patterns>

You will be shown a CUSTOMER CONDITION (a short rule over customer attributes) and a TARGET PRODUCT.
The claim under review is: "<CLAIM, e.g. customers who satisfy this condition tend not to purchase
the target product>".

The statistical evidence for this claim has already been evaluated. Do not assess whether the
pattern is statistically supported or specific to this product. Judge only whether the condition
has a plausible connection to how the target product is used.

Assign exactly one label:
  BUSINESS_RELEVANT    the condition has a plausible product-related interpretation.
  UNCERTAIN            the available domain knowledge is insufficient for a confident judgment.
  OTHER                the condition does not fit a predefined interpretation but is not invalid.
  BUSINESS_IRRELEVANT  the condition has no credible connection to the target product.

You are given no statistics and must not ask for any. Reply with JSON only:
{"label": "<LABEL>", "rationale": "<one or two sentences>"}
\end{verbatim}
\textbf{User message}
\begin{verbatim}
{
  "customer_condition": "<RULE IN PLAIN LANGUAGE, e.g. the customer had no logins in the last six months>",
  "target_product": {"name": "<PRODUCT NAME>", "function": "<PRODUCT FUNCTION>"},
  "claim_under_review": "<CLAIM>",
  "feature_definitions": {"<FEATURE>": "<TRUSTED DEFINITION>", ...}
}
\end{verbatim}
\end{minipage}
\caption{Prompt template for the LLM domain review. Bracketed fields are
placeholders filled for each rule--product pair. The review receives no
statistical scores.}
\Description{A system message that defines the reviewer role, domain knowledge,
claim, four labels, and JSON reply format, followed by a user message
containing the customer condition in plain language, the target product, the
claim, and feature definitions.}
\label{fig:llm-prompt}
\end{figure*}

\subsection{Positive Symbolic}
Positive Symbolic uses the same rules and Evidence statistic with the contrast
reversed, so the rule positive rate is compared with the category baseline.
It requires at least five training interactions in the relevant category and
rule arms, then applies only support and positive-evidence checks; $H$ and $R$
are not used.
The protection tiers are no evidence, weak ($0<E_{\mathrm{pos}}<0.20$), medium
($0.20\leq E_{\mathrm{pos}}<0.45$), and strong
($E_{\mathrm{pos}}\geq0.45$). The reported hard-tier variant samples from
these tiers in this order. It protects candidates from negative sampling and
never adds positive training examples.

\subsection{BNS and BNS-S Implementations}
Public BNS assigns item $i$ the weight
$1-\mathrm{positives}_i/\mathrm{interactions}_i$. BNS-S replaces the raw rate
with the Jeffreys posterior mean, giving
$1-(\mathrm{positives}_i+0.5)/(\mathrm{interactions}_i+1)$. Both sample $K$
distinct negatives without replacement in proportion to these weights. They
are item-level simplifications of BNS~\cite{liu2023bns}. In BNS-S + Symbolic,
the BNS-S weights are applied within each symbolic tier, and safer tiers are
exhausted before lower tiers.

\end{document}